\documentclass[letterpaper, 10 pt, conference]{ieeeconf}  

\IEEEoverridecommandlockouts                              

\usepackage{amsfonts}
\usepackage{bm}       
\usepackage{amsmath}  
\usepackage{xcolor} %
\usepackage{float}
\usepackage{subcaption}
\usepackage{graphicx}
\usepackage{booktabs} 
\usepackage[table]{xcolor} 
\usepackage{cancel}

\usepackage{array}
\usepackage{fancyhdr}
\usepackage{eso-pic} 
\usepackage{balance}

\newcolumntype{C}[1]{>{\centering\arraybackslash}p{#1}}

\title{\LARGE \bf

Event-based Motion \& Appearance Fusion for 6D Object Pose Tracking
}

\author{
Zhichao Li$^{1, 2, 3}$, Chiara Bartolozzi$^{1}$, Lorenzo Natale$^{2}$, Arren Glover$^{1}$
\thanks{$^{1}$Event-driven Perception for Robotics, Istituto Italiano di Tecnologia, Italy {\tt\small \{first.last\}@iit.it}}%
\thanks{$^{2}$Humanoid Sensing and Perception, Istituto Italiano di Tecnologia, Italy {\tt\small \{first.last\}@iit.it}}%
\thanks{$^{3}$University of Genoa, Genoa, Italy}
}

\newcommand{\firstpagecopyright}{%
    \AddToShipoutPictureFG*{%
        \AtPageUpperLeft{%
        \hspace*{\dimexpr1in+\oddsidemargin\relax}%
        \raisebox{-3.5\baselineskip}[0pt][0pt]{%
            \begin{minipage}{\textwidth}
            \centering\footnotesize
            This paper has been accepted for publication in IEEE International Conference on Robotics and Automation~(ICRA) 2026.\\
            \footnotesize{\textcopyright~2026 IEEE. Personal use of this material is permitted.
            Permission from IEEE must be obtained for all other uses, in any current or future media,
            including reprinting/republishing this material for advertising or promotional purposes,
            creating new collective works, for resale or redistribution to servers or lists, or reuse of
            any copyrighted component of this work in other works.}
            \end{minipage}
        }
        }
    }
}

\begin{document}

\firstpagecopyright

\maketitle

\fancypagestyle{withfooter}{
\renewcommand{\headrulewidth}{0pt}
\fancyfoot[C]{\footnotesize Accepted to the Challenges and Opportunities of Neuromorphic Field Robotics and Automation IEEE ICRA Workshop - 2026}
}
\thispagestyle{withfooter}
\pagestyle{withfooter}
\thispagestyle{withfooter}
\pagestyle{withfooter}

\begin{abstract}

Object pose tracking is a fundamental and essential task for robotics to perform tasks in the home and industrial settings. The most commonly used sensors to do so are RGB-D cameras, which can hit limitations in highly dynamic environments due to motion blur and frame-rate constraints. Event cameras have remarkable features such as high temporal resolution and low latency, which make them a potentially ideal vision sensors for object pose tracking at high speed. Even so, there are still only few works on 6D pose tracking with event cameras. In this work, we take advantage of the high temporal resolution and propose a method that uses both a propagation step fused with a pose correction strategy. Specifically, we use 6D object velocity obtained from event-based optical flow for pose propagation, after which, a template-based local pose correction module is utilized for pose correction. Our learning-free method has comparable performance to the state-of-the-art algorithms, and in some cases out performs them for fast-moving objects. The results indicate the potential for using event cameras in highly-dynamic scenarios where the use of deep network approaches are limited by low update rates.

\end{abstract}

\section{INTRODUCTION}

The majority of visual 6D object pose estimation literature leverages RGB and RBG-D sensors. Representative approaches include model-based techniques and template-matching methods~\cite{6696810, 7487184, 9363455, tjaden2017real, payet2011contours, hinterstoisser2010dominant}, with deep learning methods having pushed the state-of-the-art performance in recent years~\cite{hinterstoisser2012model, nguyen2022templates, lin2022single, 9811720, 9636283, wen2020se, foundationposewen2024}. Deep learning-based methods have typically higher accuracy, but they employ sophisticated network architectures and require the availability of large-scaled annotated datasets and significant computational resources for network training. Because they employ large architecture, they are demanding in terms of computation and therefore have low inference frequency, especially on resource constrained hardware. Moreover, the intrinsic limitations of RGB or RGB-D sensors, which output frames at fixed rates of 30-60 FPS, give rise to motion blur when object moves at high speeds. This effect can substantially degrade the performance of object pose trackers that depend on conventional frame-based sensors.

Event cameras are relatively novel vision sensors that have a different operating principles from the global or rolling shutters used in frame-based sensors~\cite{gallego2020event}. Event cameras measure brightness change of each pixel independently, once the pixel brightness has changed significantly an event is asynchronously output with sub-millisecon latency and precision. Due to such an operating principle, event cameras can capture object motion information with high frequency and overcome motion blur that occurs when integrating light for fixed time periods. However, correctly processing the asynchronous and discrete data encoded in events comes with its own challenges. The success of 6D object pose tracker depends critically on the effective utilization of events information and the accurate decoding motion information from events.

Only a handful of approaches have been proposed for 6D object pose tracking with event cameras. Optical flow information is almost inherently encoded in the spatio-temporal patterns formed from events and leveraging it to infer object motion presents a feasible solution~\cite{liu2024optical, li20256}. Otherwise, events have been decoded into line information, for objects primarily formed by line structures~\cite{liu2024line, liu2025stereo}, or template matching techniques have been adopted for less constrained objects~\cite{10611511}. Using a propagation and correction method has been proposed requiring a fusion with an RGB-D camera in both model-based~\cite{li20256} and data-driven neural networks~\cite{9284657}.  

\begin{figure*}[t]
    \centering
    \includegraphics[width=0.98
    \linewidth]{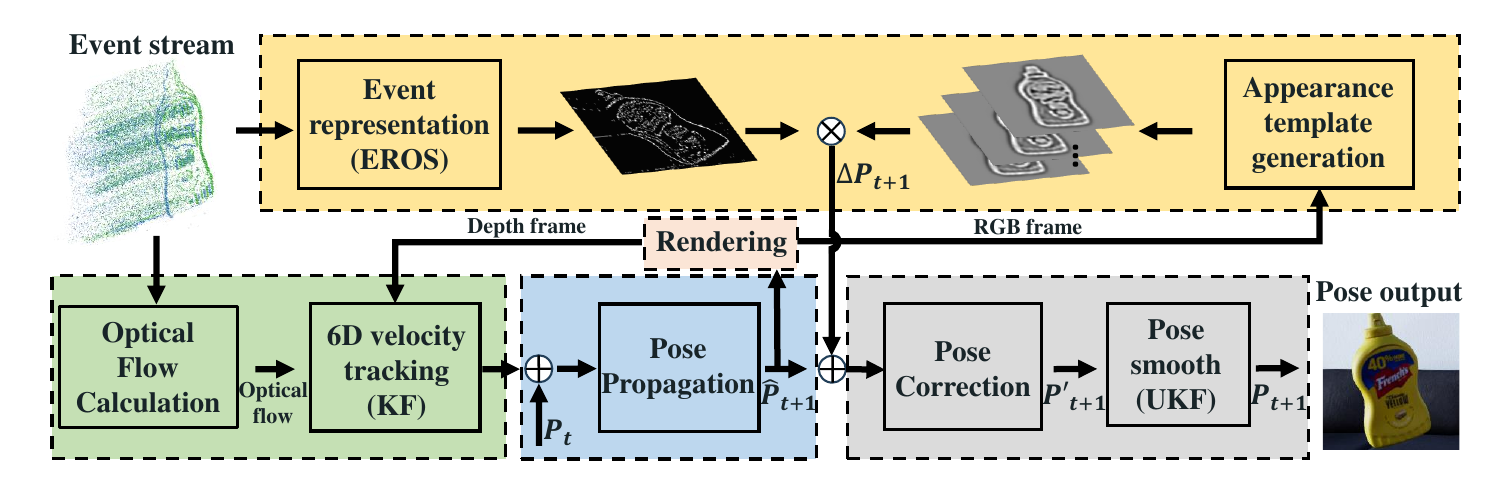}
    \caption{Overview of the proposed pipeline for 6D object pose tracking with an event camera. Events captured by an event camera are utilized to update a velocity independent representation~(EROS)~\cite{10611511}, simultaneously with event-based optical flow extraction. The 6D object pose is propagated based on the 6D object velocity estimated from the optical flow measurement. Several appearance templates are generated based on the propagated pose but with additional small pose perturbations, from which edges are extracted. The propagated 6D pose is corrected based on the best matched template compared to the EROS representation. Finally, an Unscented Kalman filter is adopted to further smooth the corrected pose over time.}   
    \label{method_pipeline}
\end{figure*}

We believe effectively leveraging event information is critical for a robust event-based 6D object pose tracker. In this work, we combine event-based optical flow together with event-based template matching to implement a high frequency 6D object pose tracker. Specifically, first we decode raw events into optical flow information which is used to predict object 6D velocity using a Kalman filter formulation. To counteract the inevitable error coming from velocity integration, a local object pose corrector is introduced. Specifically, after the pose propagation, multiple hypothesis pose templates are generated by pose perturbation using the object mesh. These hypothesis templates are compared to current sensor measurements accumulated into a spatial representation to refine the estimate directly in the object pose configuration. Our proposed method therefore reduces tracking errors accumulating over time. We decode events separately in two distinct pipelines and then organically combine them through propagation and correction. The effectiveness of this approach is validated on synthetic and real camera data in experiments.

Our contributions are the following:
\begin{itemize}
    \item We propose the Event-camera only propagation and correction method for 6D object pose tracking, that fuses an optical flow and a template matching approach.
    \item We remove the need for depth measurements in the event-driven pipeline required in 6D velocity estimation~\cite{li20256}.
    \item We perform a comparison of event-camera pose tracking against recent state-of-the-art RGB-D deep learning methods (including FoundationPose~\cite{foundationposewen2024}) obtaining comparable or better results for fast-moving object scenarios. 
\end{itemize}

\section{Related work}

Visual 6D object pose estimation involves estimating both the position and orientation of the object relative to the camera; 6D object pose tracking estimates and updates the object 6D pose over time.  Previously, probabilistic modeling methods were used for solving this task~\cite{6696810, 7487184, 9363455}.
Object pose can also be tracked, for example, by matching various templates, such as color histograms, and contours~\cite{tjaden2017real, payet2011contours, hinterstoisser2010dominant}. 

An increasing number of approaches have been proposed to address 6D object pose estimation using deep neural networks. The data-driven approaches were more effective at extracting features, enabling the generation of more efficient templates, as well as extraction of more informative key features or points for object pose tracking~\cite{hinterstoisser2012model, nguyen2022templates, lin2022single, 9811720}. Relying on a large amount of training data, deep neural networks can also learn and predict a pose variation between two continuous frames~\cite{9636283, wen2020se}, the object pose can be tracked by accumulating such pose variations over the time.  With the combination of large-scale datasets and the high performance of transformers, end-to-end modeling for object pose tracking has become viable~\cite{foundationposewen2024}. 

Approaches using RGB or RGB-D cameras can have performance degradation due to image blur for high-speed objects. In contrast, event cameras are promising sensors for 6D object pose tracking as they are largely immune to image blur, but instead lack information about color and spatial textures. One potential approach for 6D object pose tracking was to combine both RGB and event cameras using optimization and alignment techniques~\cite{9561760} or with cascading Kalman filters~\cite{10342300}.
Using only event cameras~\cite{liu2024line, liu2025stereo} propose to extract straight line features which are tracked over time. Another work, called EDOPT~\cite{10611511}, proposes a velocity-independent event representation method named EROS, which ensures consistent edge appearance for fast and slow motions. Template matching is used with EROS and object renders for pose tracking. In~\cite{10611576} a brightness increment image strategy based on motion interpolation is introduced. It can reconstruct brightness increments with events according to the object’s motion, then align the 3D object model with the incremented image directly for 6D object pose tracking.    
Among the works that are close to our work, ROFT~\cite{piga2021roft} is an optical flow aided object pose tracker. It utilizes frame-based optical flow to implement an object 6D pose velocity tracker. The object pose is propagated with 6D velocity and corrected using an deep learning based object pose estimator DOPE~\cite{tremblay2018corl:dope}. Another work adopts a similar structure, but it utilizes event-based optical flow instead of frame-based optical flow for object 6D velocity tracking~\cite{li20256}. In this work, we combine EDOPT~\cite{10611511} strategy to make local pose correction using rendered object templates, and object 6D velocity tracking from event-based optical flow~\cite{li20256}, which can provide motion information to the tracker. By using the rendered depth, the depth information is not required from cameras for 6D object velocity tracking. In this way, we build the structure of motion prediction and pose correction for 6D object pose tracking with an event camera. 

\section{Method}

\subsection{Overview}
Given the 3D model, \textcolor{black}{the initial 6D object pose} of a target object $O_{bj}$ and events stream $E:\{x_i, y_i, t_i\}$ from an event camera. Our goal is to track the 6D object pose of $O_{bj}$ with respect to the event camera's reference frame. Specifically, we want to continuously estimate 6D transformation $\bm{P}=\{\bm{t}, \bm{R}\}$ that relates $O_{bj}$ to the event camera, where $\bm{t}\in\mathbb{R}^3$ and $\bm{R}\in SO(3)$ are the translation and rotation components. 
To solve this problem, we leverage the idea of propagation and correction. Specifically, an event-based optical flow 6D object velocity tracker is utilized to propagate the object 6D motion. Then in order to eliminate pose error from pose propagation, a template-based pose correction step using events stream is executed. In the end, to smooth the tracked pose, an Unscented Kalman filter is applied before output the final pose $\bm{P}_{t+1}$. The pipeline is shown in Fig.~\ref{method_pipeline}.

\subsection{Event optical flow based velocity tracking}

\subsubsection{Event-based optical flow}

In this work, the event-based optical flow is calculated by examining spatio-temporal relationship for each event within a region of interest (\textit{RoI}). Then apply a registration technique to obtain the event-based optical flow~\cite{li20256}. Specifically, when an object moves in 3D space, one single point on the object crosses the 2D image plane and triggers events. Assume three events $e_i=\{\bm{x}_i, t_i\}$, $e_j$ and $e_k$ are triggered by the motion of the same point and in a very short time period. They follow spatio-temporal constraint:  
\begin{equation}
    \begin{split}
        \frac{\bm{x}_j - \bm{x}_i}{t_j-t_i}  = \frac{\bm{x}_k - \bm{x}_j}{t_k-t_j + \tau}\\
    \end{split}
\end{equation}%
 $\bm{x}_i$, $\bm{x}_j$ and $\bm{x}_k$ are 2D pixel coordinates of triggered events $e_i$, $e_j$ and $e_k$. $\tau$ is the parameter that represents the tolerance in time. These three events are spatio-temporal triplet matched events.
The optical flow at pixel coordinate $\bm{x}_i$ can be calculated as:
\begin{equation}\label{eq:flow_calculation}
        \bm{F}(\bm{x}_i) = (\bm{x}_k - \bm{x}_i)/(t_k-t_i)
\end{equation}
However, single triplet matched events is unable to guarantee events $e_i$, $e_j$ and $e_k$ are triggered by the motion of the same object point. Incorrect triplet matched events \textcolor{black}{from wrong matching, background noise and sensor noise} will introduce errors in the optical flow computation. To eliminate this error, we follow the work~\cite{li20256} using spatio-temporal registration strategy to improve the event optical flow from triplet matches. By splitting the image space into small grids. Search triplet matched events within each grid if events 
are triggered. Each grid is defined as a \textit{RoI}.
In each \textit{RoI}, all searched triplet events are used for estimating a single optical flow vector $\bm{F}$ using the following cost function:
\begin{equation}\label{eq:least-square}
        \bm{F} = \arg\min_{\bm{F} \in \mathcal{M}}\bigg(\sum\limits_{i=1}^{n}\min_{\bm{F}_{ic}\in M_i}\mid \bm{F}-\bm{F}_{ic}\mid \bigg)
\end{equation}%
$\bm{F}$ denotes the target optical flow, which minimizes the cost function.  $\mathcal{M}$ represents the set of candidate flow vectors computed from the identified triplets using Eq.~\ref{eq:flow_calculation}, within a region of interest (\textit{RoI}) of $n$ events. For a single event $\boldsymbol{e}_i$, the subset $M_i \subset \mathcal{M}$ contains all candidate flow vectors associated with $\boldsymbol{e}_i$. \textcolor{black}{With the \textit{RoI} and perform spatio-temporal registration more candidate triplet matches are incorporated, thereby enabling effective suppression of background and the sensor-induced noise}.

\subsubsection{6D Object velocity estimation with optical flow}
Given event stream $E:\{x_i, y_i, t_i\}$ after decoded them as event optical flow $\bm{F}_i$. We estimate the 6D object velocity $\bm{\mathcal{V}}_t$ from $\bm{F}_i$ using a Kalman filter as~\cite{li20256}.
\begin{equation}\label{eq:ekf_state}
    \begin{split}
        \bm{\mathcal{V}}_t=[\bm{v}_{ot}, \bm{\omega}_{ot}]
    \end{split}
\end{equation}
Where, $\bm{v}_{ot} \in \mathbb{R}^3$ denotes the linear velocity of a point that momentarily coincides with the camera's origin and moves as though it were rigidly attached to the object. $\bm{\omega}_{ot} \in \mathbb{R}^3$ represents the object's angular velocity relative to that same point.
 Similarly to~\cite{li20256} a motion model for the Kalman filter is defined:
\begin{equation}\label{eq:ekf_motion_model}
	\begin{split}
		\bm{\mathcal{V}}_{t+1} &= \rho \bm{\mathcal{V}}_{t} + \bm{w},
	\end{split}
\end{equation}
where the propagation noise, $\bm{w} \sim N\left(0, \mathrm{diag}(\bm{Q}_{v}, \bm{Q}_{w})\right)$, follows a zero-mean Gaussian distribution with covariance $\bm{Q}_{v} \in \mathbb{R}^{3\times 3}$ and $\bm{Q}_{w} \in \mathbb{R}^{3\times 3}$ associated to the linear and angular velocity, respectively. $\rho$ is a decay parameter, following~\cite{li20256} we set it to 0.5.

The measurement model of Kalman filter is built as:
\begin{equation}\label{eq:ekf_measurement_model}
    \bm{y} = (\bm{F}(u,v) - \bm{\hat{F}}(u,v))\textcolor{black}{\Delta_{T}}
\end{equation}
where $\bm{F}(u,v)$ is event optical flow at pixel coordinate $(u,v)$ and $\bm{\hat{F}}(u,v) = \bm{\mathcal{J}}(u, v)\bm{\mathcal{V}}_t$, represents the mapped 2D pixel velocity from estimated 6D object velocity using the interaction matrix $\bm{\mathcal{J}}(u, v)$~\cite{li20256}. \textcolor{black}{$\Delta_{T}$ represents the time interval between two consecutive updates of the object velocity tracker.} Different from work~\cite{li20256}, which uses depth captured from the camera, in this work, depth is obtained by rendering using the tracked 6D object pose.

\subsubsection{6D Object pose propagation} 
Given the tracked 6D object velocity $\bm{\mathcal{V}}_t(\bm{v}_{ot}, \bm{\omega}_{ot})$ from Kalman filter, the 6D object pose $\bm{P}_t=\{\bm{t}_t, \bm{R}_t\}$ can be propagated as $\hat{\bm{P}}_{t+1}$. \textcolor{black}{$\bm{\mathcal{V}}_t(\bm{v}_{ot}, \bm{\omega}_{ot})$ is splitted into translation and rotation components and used in the Eq.~\ref{eq:pose_propgation_vel} separately}.
\begin{equation}\label{eq:pose_propgation_vel}
\begin{aligned}
    &\hat{\bm{{t}}}_{t+1} = \bm{{t}}_{t} + \bm{v}_{ot}\cdot\Delta{t}\\
    \hat{\bm{{q}}}_{t+1} = cos(|&|\bm{\omega}_{ot}\frac{\Delta{t}}{2}||)\bm{I}_4 + \frac{sin(||\bm{\omega}_{ot}\frac{\Delta{t}}{2}||)}{||\bm{\omega}_{ot}||}\Omega(\bm{{q}}_t)
\end{aligned}
\end{equation}
For propagating the rotation part, we process it in quaternion format. $\bm{q}_t$ is quaternion format of the rotation component $\bm{R}_t$. And $\Omega(\bm{{q}}_t)$ follows~\cite{piga2022hybrid}:
\begin{equation}\label{eq:skew-symmetric matrix}
    \Omega(\bm{q}_t) =
    \begin{bmatrix}
    -\omega_x && -\omega_y && -\omega_y && 0\\
    \omega_x  &&     0     && -\omega_z && \omega_y\\
    \omega_y  && \omega_z  &&  0        &&-\omega_x\\
    \omega_z  && -\omega_y && \omega_x  && 0
     \end{bmatrix} 
\end{equation}

\subsection{Local pose correction}
Following Eq.~\ref{eq:pose_propgation_vel}, inaccuracies in the estimated 6D object velocity $\bm{\mathcal{V}}_t$ lead to errors in the tracked pose. Over time, these errors accumulate, eventually leading to tracker failures. To solve this problem we introduce a  local correction strategy. After propagating the 6D object pose using the estimated velocity, the 6D object pose is corrected by a template-based local pose corrector, inspired from~\cite{10611511}.


\begin{figure}[t]
  \centering
  \begin{subfigure}[t]{0.3\linewidth}
    \centering
    \includegraphics[width=\linewidth]{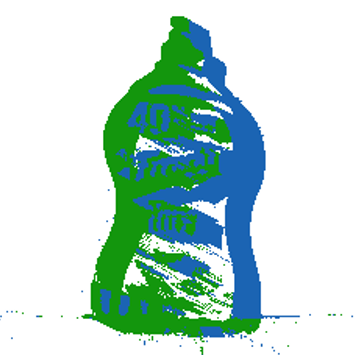}
    \caption{}
    \label{fig:raw}
  \end{subfigure}
  \hspace{0.02\linewidth} 
  \begin{subfigure}[t]{0.3\linewidth}
    \centering
    \includegraphics[width=\linewidth]{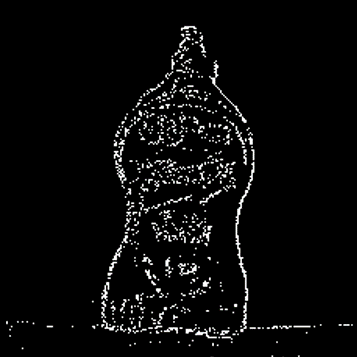}
    \caption{}
    \label{fig:processed}
  \end{subfigure}
  \hspace{0.02\linewidth} 
  \begin{subfigure}[t]{0.3\linewidth}
    \centering
    \includegraphics[width=\linewidth]{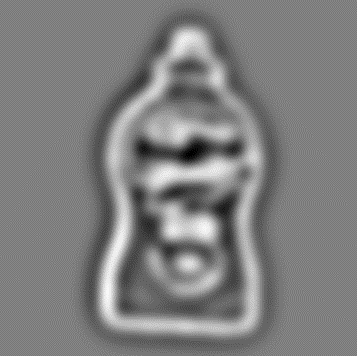}
    \caption{}
    \label{fig:expectation}
  \end{subfigure}
  \caption{Visualization of (a) raw events, (b) EROS representation, and (c) template generated using the model rendered at a perpetuated pose with edge gradient extraction.}
  \label{fig:comparison}
\end{figure}


\subsubsection{Event representation}
To allow for efficiently extracting object appearance, including outlines and edges, whenever the object is moving, the asynchronous stream is converted to EROS~\cite{10611511}. Different from raw events (Fig.~\ref{fig:raw}), EROS creates an image-like representation where each pixel encodes the presence of a light change (see Fig.~\ref{fig:processed}). This velocity-independent representation is updated asynchronously, with each event producing a small, localized adjustment. Specifically, we define an array $\mathcal{E}$ with the same size as the image plane, where each element takes a value in the range $[0.0, 1.0]$.
When an new event is triggered at $(u, v)$, we set $\mathcal{E}(u, v) = 1.0$, and decay the values of all neighboring pixels in the region $\mathcal{N}(u, v)$ by a factor $\lambda$. 
\begin{equation}\label{eq:eros_neighbor}
    \begin{split}
    \mathcal{E}(x, y) =& \lambda \times \mathcal{E}(x, y) \\
    \forall(x, y) &\in \mathcal{N}(u, v)
    \end{split}
\end{equation}
$(x, y)$ represents the pixel within the neighborhood of pixel $(u, v)$.
\subsubsection{Gradient visual expectation}
We generate the expected object image gradients for the propagated object pose $\hat{\bm{
P}}_{t+1}$. By projecting the known 3D mesh onto the 2D image plane using the camera's intrinsic parameters, we render a synthetic image that represents how the object should appear from that viewpoint. This projection simulates the object's outline and surface details as seen by the camera. Next, to extract meaningful feature from this synthetic image, following~\cite{10611511} we apply a Sobel filter, which highlights the edges and gradients-areas where the image intensity changes sharply and could trigger events~(see Fig.~\ref{fig:expectation}). The resulting expectation map serves as an appearance template against the observation in event representation of the object.
A Difference of Gaussians filter is applied to broaden edges, allowing some matching tolerance, produce a gradient to differentiate strong and weak matches, and to produce a negative matching region to reject badly matching observations. 

\subsubsection{Template-based pose correction}
We aim to predict how the current object pose could have evolved such that it would generate the events observed by the event camera. Since an event camera has no integration period we assume small pose changes between two consecutive estimations. Specifically, we assume that the templates resulting from the projection of two consecutive poses have no more than one pixel of displacement. 

The relationship between changes in the object pose $\Delta\{x, y, z, \alpha, \beta, \gamma\}$ (specific for translations) and changes in the image plane $\Delta\{u, v\}$ (pixel motion) can be captured by the image Jacobian, which describes how each degree of freedom (move forward/backward, left/right, up/down along the axes) translates into pixel shifts in the image plane. 

\begin{equation}\label{eq:image_jacobian}
   \begin{bmatrix}
    \Delta{u}\\\Delta{v}
    \end{bmatrix}
   =
    \begin{bmatrix}
    -\frac{f_x}{z} & 0 & \frac{u-c_x}{z} \\
    0 & -\frac{f_y}{z} & \frac{v-c_y}{z} 
    \end{bmatrix} 
    \begin{bmatrix} 
    \Delta{x} & \Delta{y} & \Delta{z}
    \end{bmatrix}^T
\end{equation}
We perturb $\Delta\bm{P}_{t+1}$ for the propagated object pose $\bm{\hat{P}}_{t+1}$ such that the resulting motion on the image plane is at $\Delta{p}$ pixels (in this work $\Delta{p}$ set as one pixel). 
By inverting the Jacobian and decoupling each dimension, we derive the amount of translation along x or y axis for pose corresponds to $\Delta{p}$ pixel image motion:
\begin{equation}\label{eq:pose_perturbation_translation}
    \left\{ 
        \pm \Delta x = \frac{\mp z \Delta{p}}{f_x} \quad 
        \pm \Delta y = \frac{\mp z \Delta{p}}{f_y} \quad 
        \pm \Delta z = \frac{\pm z\Delta{p}}{\bar{p}} 
    \right\}
\end{equation}
$\bar{p}=||(\Delta{u}, \Delta{v})||$. For rotations, assume the object rotates with a small rotation degree $\theta$ (in this work $\theta$ is set as ${0.5}^\circ$). Then
\begin{equation}\label{eq:pose_perturbation_rotation}
    \left\{ 
        \pm \Delta \alpha = \pm \theta \quad 
        \pm \Delta \beta = \pm \theta \quad 
        \pm \Delta \gamma = \pm \theta 
    \right\}
\end{equation}
These equations generate 12 perturbed poses (positive and negative directions along each motion axis). If considering an additional null motion hypothesis, these give a total 13 hypothesis poses. The hypothesis pose is defined as $\{H_i|i\in\left[1,...,13\right]\}$. 

To reduce the pose error accumulated by pose propagation in eq.\ref{eq:pose_propgation_vel}. 13 hypothesis pose $H_i$ is generated based on propagated object pose $\hat{\bm{{P}}}_{t+1}$, and compare them with current object pose observation EROS $O_{t+1}$ from the camera. The hypothesis pose $H_i$ with the highest similarity with the observation $O_{t+1}$ is selected and the pose $\hat{\boldsymbol{P}}_{t+1}$ is refined as:
\begin{equation}\label{eq:pose_perturbation_update}
    \boldsymbol{P'}_{t+1} = \hat{\bm{P}}_{t+1} + \operatorname*{argmax}_{\Delta \bm{P}_{t+1}} \{H_i \cdot O_{t+1}\}, i \in[1,...,13].
\end{equation}

\begin{figure}[t]
    \begin{subfigure}[t]{0.5\linewidth}
        \centering
        \includegraphics[width=0.8\textwidth,trim={0 0 7.8cm 0},clip]{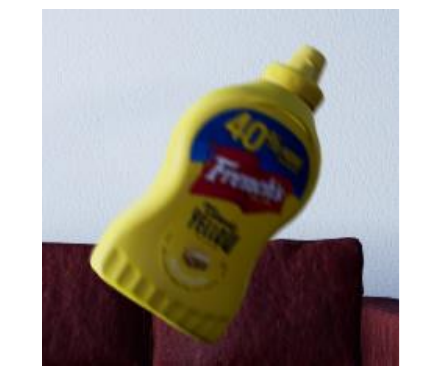}
        \caption{} \label{fig:regularmotion}
    \end{subfigure}%
    \begin{subfigure}[t]{0.5\linewidth}
        \centering
        \includegraphics[width=0.8\textwidth,trim={7.8cm 0 0 0},clip]{motion_blur.pdf}
        \caption{} \label{fig:fastmotion}
    \end{subfigure}
    \caption{Snapshots of objects in synthetic data sequences: (a) regular motion sample, and (b) fast motion sample.}   
    \label{motion_blur}
\end{figure}

\subsection{Pose smooth with UKF}
We employ an Unscented Kalman Filter (UKF) to
smooth the refined object pose $\bm{P'}_{t+1}$ and output the final pose $\bm{P}_{t+1}$ . The UKF effectively handles the non-nonlinearities inherent in the pose dynamics by propagating a set of sigma points through the system model. It improves the temporal consistency of the refined pose while reducing the effects of perturbations in the appearance template pose correction, as well as mitigating pose observation noise from the camera.

\begin{table*}[pt!]
    \vspace{0.7em}
    \renewcommand\tabcolsep{2.3pt}
	\centering
    \scriptsize
    \begin{tabular}{c c|c c| c c| c c| c c}
		\cline{1-8} 
        \toprule[1.1pt]
		\multicolumn{2}{c}{\textbf{Sequence}}                                        
		&\multicolumn{2}{c}{\textbf{mustard\_regular}}  
        &\multicolumn{2}{c}{\textbf{potted\_regular}}  
        &\multicolumn{2}{c}{\textbf{cracker\_regular}}  
        &\multicolumn{2}{c}{\textbf{tomato\_regular}}  
        \\
		\hline
        Methods  
        & Modality
        & $e_{p}(\sigma)(cm)$     
        & $e_{r}(\sigma)(deg)$   
        & $e_{p}(\sigma)(cm)$     
        & $e_{r}(\sigma)(deg)$ 
        & $e_{p}(\sigma)(cm)$     
        & $e_{r}(\sigma)(deg)$   
        & $e_{p}(\sigma)(cm)$     
        & $e_{r}(\sigma)(deg)$
        \\
		\hline

        \multicolumn{1}{c|}{ROFT~\cite{piga2021roft}}
        & RGBD
        &1.16(0.59)   &4.88(1.62)
        &3.41(1.44)   &19.81(8.78)
        &1.54(0.40)   &\cellcolor[gray]{0.90}{2.61}(1.04) 
        &\cellcolor[gray]{0.90}{1.77}(0.63)   &\cellcolor[gray]{0.90}{9.29}(3.87)\\

       \multicolumn{1}{c|}{se(3)-TrackNet~\cite{wen2020se}}
        & RGBD
	    &\cellcolor[gray]{0.75}{0.77}(0.09)  &\cellcolor[gray]{0.75}{2.78}(1.11)  
        &\cellcolor[gray]{0.90}{0.84}(0.34)   &\cellcolor[gray]{0.90}{11.37}(5.81)
        &\cellcolor[gray]{0.90}{0.86}(0.52)   &5.20(3.32) 
        &28.27(8.81)  &34.90(9.41)\\

        \multicolumn{1}{c|}{FoundationPose~\cite{foundationposewen2024}}
        & RGBD
		&\cellcolor[gray]{0.90}{0.86}(0.40)   &\cellcolor[gray]{0.90}{6.29}(3.36)  
        &\cellcolor[gray]{0.75}{0.56}(0.18)
        &\cellcolor[gray]{0.75}{1.72}(0.75)
        &\cellcolor[gray]{0.75}{0.40}(0.15) 
        &\cellcolor[gray]{0.75}{0.94}(0.35)   &\cellcolor[gray]{0.75}{0.48}(0.21)  
        &\cellcolor[gray]{0.75}{1.53}(0.70)\\ 

        \multicolumn{1}{c|}{Hybrid method~\cite{li20256}}
        & RGBDE
		&1.61(0.68)   &5.78(1.75)  
        &4.42(3.31)   &12.96(7.99)
        &1.95(1.09)   &7.30(3.66)   
        &2.13(0.84)   &9.25(4.08)\\

        \multicolumn{1}{c|}{EDOPT~\cite{10611511}}  
        & Events
		&1.93(1.04)    &14.17(8.40)
        &15.55(8.90)   &21.89(10.57) 
        &1.10(0.59)    &4.11(2.31) 
        &13.15(7.33)   &71.96(36.60)\\

        \multicolumn{1}{c|}{EDOPT\_Vel}   
        & Events
		&10.04(7.08)   &29.92(15.60)
        &9.69(6.51)    &17.56(9.44) 
        &1.26(0.66)    &3.49(1.44) 
        &25.65(8.80)   &68.74(19.63)\\ 

        \multicolumn{1}{c|}{Ours~(only Vel)}   
        & Events
		&8.34(2.80)    &25.87(12.48)
        &7.77(3.13)    &14.01(4.38) 
        &5.73(2.02)    &20.82(6.40) 
        &4.01(1.12)    &21.62(4.76)\\ 
        
		\multicolumn{1}{c|}{Ours (w/o ukf)} 
        & Events
		&1.79(1.03)  &9.53(5.90)
        &1.68(0.97)  &11.92(6.68)
        &0.98(0.49)  &3.01(1.29)  
        &8.26(3.87)  &61.15(29.76) \\

		\multicolumn{1}{c|}{Ours (ukf)}        
		& Events
        &1.66(0.94)  &7.84(4.40)
        &1.90(0.80)  &14.14(6.65)  
        &1.13(0.61)  &3.32(1.50) 
        &7.66(3.10)  &58.36(24.38) 
        \\ 
        
        \hline

		\multicolumn{2}{c}{\textbf{Sequence}}                                        
		&\multicolumn{2}{c}{\textbf{mustard\_fast}}  
        &\multicolumn{2}{c}{\textbf{potted\_fast}}  
        &\multicolumn{2}{c}{\textbf{cracker\_fast}}  
        &\multicolumn{2}{c}{\textbf{tomato\_fast}}  
        \\

		\hline
        \multicolumn{1}{c|}{ROFT~\cite{piga2021roft}}
        & RGBD
        &4.95(2.64)    &21.49(10.95)  
        &74.12(35.59)  &80.28(38.73)
        &6.76(3.21)    &27.73(17.51) 
        &3.25(1.41)    &17.40(7.58)\\

        \multicolumn{1}{c|}{se(3)-TrackNet~\cite{wen2020se}}
        & RGBD
	    &89.82(47.29)  &87.64(49.20)   
        &15.54(6.37)   &16.24(6.35)
        &24.13(9.62)   &29.90(11.86) 
        &26.04(9.67)   &22.48(8.40)\\

        \multicolumn{1}{c|}{FoundationPose~\cite{foundationposewen2024}}
        & RGBD
		&\cellcolor[gray]{0.90}{1.49(0.67)}   &12.69(7.36)  
        &\cellcolor[gray]{0.75}{1.04(0.44)}   &\cellcolor[gray]{0.75}{6.33(3.99)} 
        &1.42(0.57)  &3.70(2.61)   
        &\cellcolor[gray]{0.75}{0.89(0.50)}   &\cellcolor[gray]{0.75}{4.51(3.11)}\\

        \multicolumn{1}{c|}{Hybrid method~\cite{li20256}}
        & RGBDE
		&4.66(2.29)   &\cellcolor[gray]{0.90}{11.90}(4.63)  
        &4.90(2.13)   &14.82(6.09)
        &3.30(1.93)   &9.13(4.96)   
        &2.89(1.27)   &10.38(4.35)\\ 
        
        \multicolumn{1}{c|}{EDOPT~\cite{10611511}}  
        & Events
		&26.31(9.45)   &46.77(22.41)
        &12.03(5.79)   &20.03(8.58) 
        &24.78(8.99)   &33.95(13.30)  
        &19.25(11.74)  &31.79(18.11)\\

        \multicolumn{1}{c|}{EDOPT\_Vel}   
        & Events
		&38.24(13.18)   &82.40(30.43)
        &2.22(1.01)     &12.12(5.34) 
        &33.76(14.72)   &27.88(11.97)  
        &75.64(53.78)   &42.25(21.18)\\ 

        \multicolumn{1}{c|}{Ours~(only Vel)}   
        & Events
		&11.18(5.34)   &13.23(5.56)
        &12.77(5.76)   &15.67(6.88) 
        &16.82(7.58)   &24.43(13.28)  
        &7.60(2.57)    &24.67(13.69)\\ 
        
		\multicolumn{1}{c|}{Ours(w/o ukf)} 
        & Events
		&2.54(1.71)  &15.50(9.27)
        &1.59(0.70)  &14.12(6.53)
        &\cellcolor[gray]{0.90}1.12(0.54)  &\cellcolor[gray]{0.90}3.06(1.29)  
        &\cellcolor[gray]{0.90}{1.51(0.74)}  &5.68(2.20) \\

		\multicolumn{1}{c|}{Ours(ukf)}        
		& Events
        &\cellcolor[gray]{0.75}{1.14}(0.57)  &\cellcolor[gray]{0.75}{11.53}(5.68)
        &\cellcolor[gray]{0.90}1.30(0.64)    &\cellcolor[gray]{0.90}13.61(6.44)  
        &\cellcolor[gray]{0.75}{0.90}(0.47)  &\cellcolor[gray]{0.75}{3.05}(1.23)  
        &1.74(0.84)    &\cellcolor[gray]{0.90}4.68(1.89) 
        \\ 
	
        \bottomrule[1.1pt]

	\end{tabular}
    \caption{6D object pose tracking performance of multiple algorithms using different information modalities on synthetic object motion sequences. The results of translations and rotations are measured by root mean square error~(RMSE) and standard deviation~($\sigma$). EDOPT\_Vel means that the EDOPT algorithm integrates the velocity calculated from two recent continuously tracked object poses, and Ours~(only Vel) refers to propagating the pose using only 6D velocity.}
    \label{Main_table}
\end{table*}

\section{Experiment}
To evaluate the performance of our proposed method, we execute experiments on synthetic and real world data from event cameras and compare with multiple algorithms including RGB/RGBD frame-based, events-based and hybrid methods. \textcolor{black}{We assume a static camera setting, and the object in the scene is moving. Under this assumption, cluttered environments with background motion are uncommon, as static background structures do not generate events and do not interfere with the proposed method.} In the experiments, we set both event-based optical flow update frequency and event representation EROS update rate at 500Hz. 

\subsection{Testing data}
\begin{figure}[t]
    \centering
    \includegraphics[width=0.55
    \linewidth]{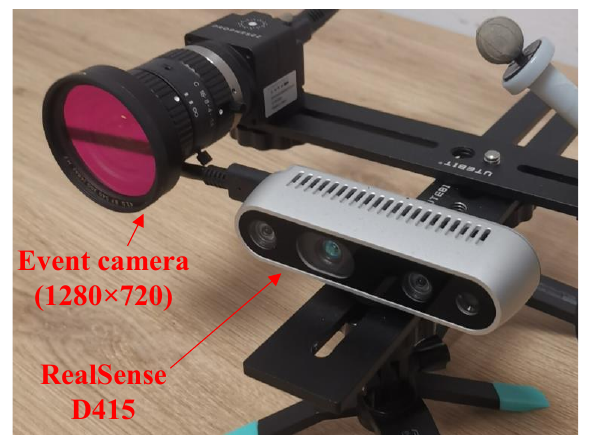}
    \caption{Dual-camera setup. \textcolor{black}{The left device is an event camera with resolution 1280$\times$720. The right RGB-D camera is a RealSense D415, only used for evaluating base-line algorithms and doesn't form an input to our algorithm.} }   
    \label{Dual-camera setup}
\end{figure}

Two datasets sequences are evaluated in the experiments. The first is a synthetic dataset from~\cite{li20256}. It is generated using a simulation engine, while the second dataset is acquired with a dual-camera setup, consisting of an event camera (1280$\times$720 resolution) and an Intel RealSense D415, shown in Fig.~\ref{Dual-camera setup}. Several objects from YCB object sets, such as cracker\_box, mustard\_bottle, potted\_meat\_can, etc, are used for data collection.  

For synthetic data sequences, the data includes triggered events, RGB-D frames recorded at 500 FPS, and corresponding ground-truth object poses also provided at 500 Hz. To simulate the frames with motion blurs as collected from a real camera, frames are averaged and achieves 60~FPS output, Fig.~\ref{motion_blur}. Moreover, in order to simulate the real depth captured by the RealSense, a Gaussian noise with 25~mm standard deviation is added. For real-world data sequences, objects are manipulated by a human operator. The data contains events induced by object motions, as well as RGB and depth frames captured at 60 FPS by the RealSense. In these data sequences, ground-truth object poses are unavailable.

\subsection{Metrics}\label{sec:metrics}
In the experiment, the tracking performance for synthetic data sequences is evaluated quantitatively using root mean square error (RMSE) with respect to the ground-truth pose. Positional errors are quantified with the Euclidean distance, whereas rotational errors are expressed as angular vectors. The standard deviation ($\sigma$) is also calculated for evaluating the smoothness of the tracking trajectories. 
For the real-world sequences, qualitative comparison results are reported. The estimated poses are examined by rendering the object according to the tracked pose and overlaying it onto the captured events stream or RGB frames.  

\subsection{Comparison against RGB/RGBD frame-based model}
We evaluate the performance of our algorithm and compared it with several frame-based algorithms including ROFT~\cite{piga2021roft}, se(3)-TrackNet~\cite{wen2020se} and FoundationPose~\cite{foundationposewen2024} on synthetic data sequences. ROFT is frame-based optical flow aided method with dual filter structure, it combines the 6D object velocity calculated from optical flow with the pose estimated by the learning-based pose estimator DOPE~\cite{tremblay2018corl:dope}. While se(3)-TrackNet and FoundationPose are both learning based methods. The object motion sequences can be split into \textit{regular} and \textit{fast} groups by pixel motion speeds. From Table~\ref{Main_table}, algorithms that use RGBD modality outperform our method in \textit{regular} motion speed group. When the motion speed is not very high, the captured frames show little or almost no noticeable motion blur, higher tracking accuracies demonstrate the effectiveness of learning-based methods. However, in the \textit{fast} motion group, high speed motions cause significant frame blurs, which reduces the accuracy of frame-based optical flow and also affects the effectiveness of neural network-based feature extractors. In the end, it degrades the performance of ROFT and se(3)-TrackNet. FoundationPose adopts a transformer architecture, which has stronger and stable feature extraction capability. Therefore, the impact of the high speed motion on the performance less significant in this experiment. The advantage of our method becomes clear in the high speed motion group and achieves comparable results to FoundationPose. This is because high speed helps trigger a sufficient number of events for optical flow extraction, which further improves the accuracy of 6D object velocity tracker.        

\subsection{Comparison against hybrid method}
We also compared against a hybrid method~\cite{li20256}, which employs both events and frames: it tracks 6D object velocity from event optical flow and fuses it with a learning-based object pose estimator DOPE~\cite{tremblay2018corl:dope}, the same algorithm used in ROFT~\cite{piga2021roft}. Our method outperforms this hybrid method in tracking accuracy, except for the tomato sequence with normal motion speed. 

This performance difference can be attributed to the local pose correction in our method and the global object pose estimator DOPE. Our local pose correction relies on the velocity-independent representation, it will not be affected by motion blur. In contrast, DOPE relies on RGB frames, which are susceptible to motion blur. This comparison demonstrates the effectiveness of our local pose correction strategy. 

\subsection{Comparison against Event-based model}
Finally, we compared our proposed method with the event-based method EDOPT~\cite{10611511}. EDOPT relies on the velocity-independent representation EROS and produces templates by applying perturbations. In comparison with our method, EDOPT relies solely on a local correction strategy, \textcolor{black}{ without incorporating a motion model from motion cues. When the object moves at higher speed, the local pose correction space may fail to cover the ground truth pose, leading to tracking failure. As shown in Table \ref{Main_table}, EDOPT tracks the \textit{mustard\_regular} and \textit{cracker\_regular} sequences, but fails the remaining sequences. In contrast, our method achieves superior tracking performance in terms of tracking accuracy across all evaluated sequences.}


\begin{figure}[t]
  \centering
    \includegraphics[width=\linewidth]{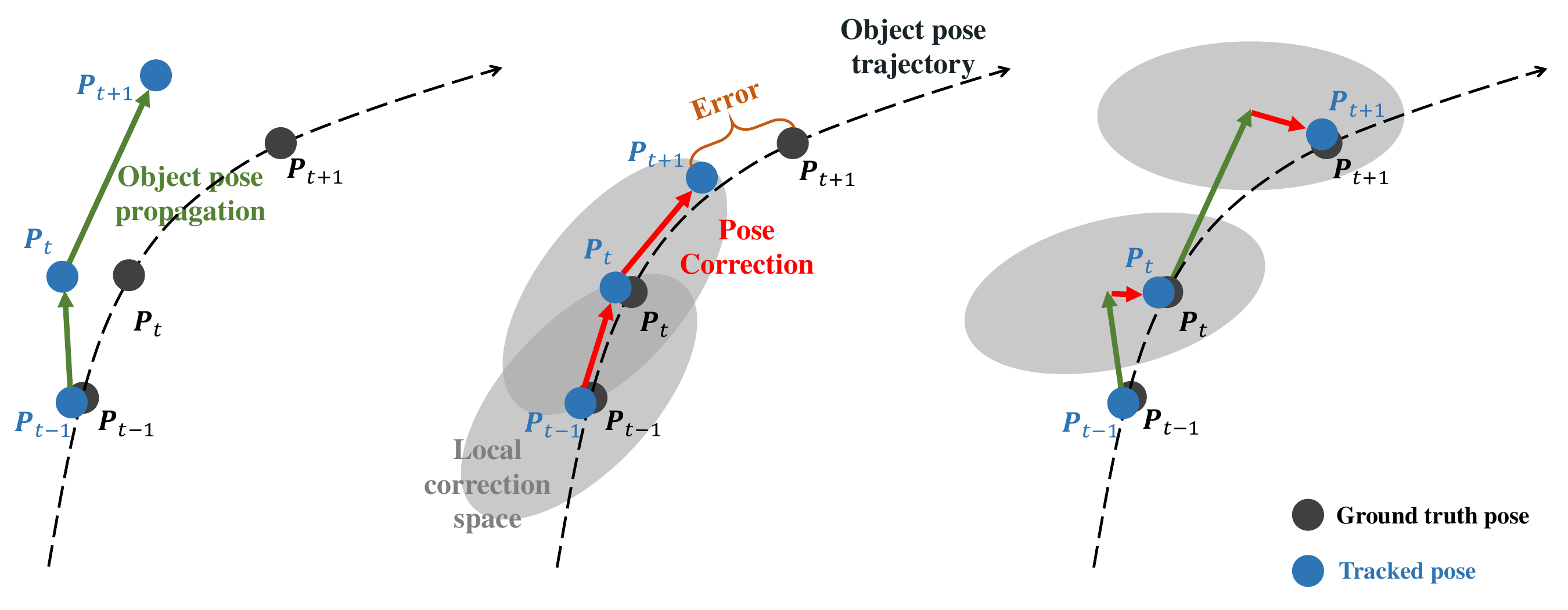}
    \caption{Propagation and local correction strategy. The (left) figure shows error accumulation if velocity integration is used alone. The (middle) figure demonstrates a failure case of perturbation correction, when the noise or discretisation error allow the true pose to leave the perturbation region without correction. The (right) figure shows the proposed method combining both approaches, enabling the pose variation between the propagated pose and the ground truth pose to more likely remain within the bounds of the local correction space, and conversely correct integration error.}
    \label{fig:pose_refinement}

\end{figure}

\begin{figure*}[t]
    \centering
    \includegraphics[width=1
    \linewidth]{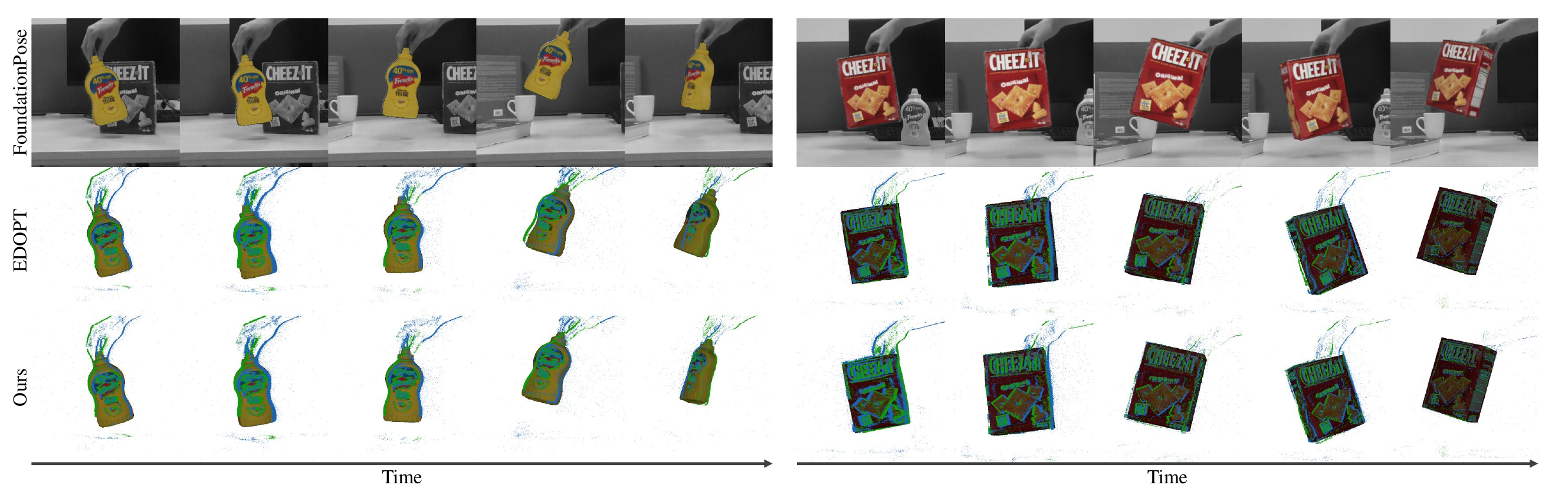}
    \caption{Qualitative results of mustard\_bottle and  cracker\_box  from real camera data.}   
    \label{real_camera_data_testing_mustard}
    
\end{figure*}

\subsection{Real camera data testing}
In the real camera data testing, we compare our method with EDOPT~\cite{10611511} and FoundationPose~\cite{foundationposewen2024}. As the testing sequences do not provide ground truth object poses, we report only qualitative results for the evaluated algorithms. In particular, the tracked object pose is visualized by rendering the object using its mesh onto the collected events streams for event-based algorithms and onto the RGB frames. The performance of the algorithms is evaluated by assessing how well the rendered object aligns with the events streams over a short temporal window or with individual RGB frames.

As shown in Fig.~\ref{real_camera_data_testing_mustard}, the rendered objects align well with the event streams for our algorithm and with the RGB frames for FoundationPose over time. The performance of the two algorithms is comparable on the testing sequences.
In contrast, the rendered object aligns well with events stream at the beginning, but misalignments become evident over time for EDOPT. Those results are consistent with the findings from the experiments on synthetic data sequences. 

\subsection{Ablation study}
Our method integrates object velocity from event-based optical flow with a Kalman filter and a local pose correction module based on templates. We conduct ablation studies on synthetic data sequences to assess components of our proposed method. All results are obtained by averaging over all sequences and reported in terms of RMSE and $\sigma$ as mentioned in Sec.~\ref{sec:metrics}.

\subsubsection{Integration of 6D object velocity and local pose correction}

\begin{table}[t]
    \centering
    \label{tab:ablation}
    
    \newcommand{\Wmethod}{4.4cm}
    \newcommand{\Wval}{1.5cm}
    
    \begin{subtable}[t]{\linewidth}
        \centering
        \caption{Tracking performance with different modules.}
        \label{Table_ablation:6d_object_velocity_integration}

        \begin{tabular}{@{}C{\Wmethod} c c}
        \toprule[1.1pt]
        Method & $e_{p}(\sigma)$ (cm) & $e_{r}(\sigma)$ (deg) \\
        \midrule
        Local correction     & 14.26(6.73) & 30.58(15.04) \\
        Velocity propagation & 9.28(3.79)  & 20.04(8.43)  \\
        Ours                 &\cellcolor[gray]{0.75}\;\;2.18(1.00)\;\;  &\cellcolor[gray]{0.75}\;14.57(6.52)\; \\ 
        \bottomrule
        \end{tabular}
    \end{subtable}
    
    \vspace{0.8em}
    
    \begin{subtable}[t]{\linewidth}
        \centering
        \caption{Tracking performance using different motion information.}
        \label{Table_ablation:event_op_velocity}
        
        \begin{tabular}{@{}C{\Wmethod} c c}
        \toprule
        Method & $e_{p}(\sigma)$ (cm) & $e_{r}(\sigma)$ (deg) \\
        \midrule
        Local correction 6D object velocity & 24.56(13.22) & 35.55(14.38) \\
        Optical flow based 6D object velocity &\cellcolor[gray]{0.75}2.18(1.00)  &\cellcolor[gray]{0.75}14.57(6.52) \\ 
        \bottomrule
        \end{tabular}
    \end{subtable}
    
    \vspace{0.8em}
    
    \begin{subtable}[t]{\linewidth}
        \centering
        \caption{Performance improvement achieved by introducing UKF.}
        \label{Table_ablation:integrate_ukf}
        
        \begin{tabular}{@{}C{\Wmethod} c c}
        \toprule
        Method & $e_{p}(\sigma)$ (cm) & $e_{r}(\sigma)$ (deg) \\
        \midrule
        w/o UKF & \;\;2.43(1.26)\;\; &\;15.50(7.87)\;  \\
        w UKF &\cellcolor[gray]{0.75}2.18(1.00)  &\cellcolor[gray]{0.75}14.57(6.52) \\ 
        \bottomrule
        \end{tabular}
    \end{subtable}
    
    \caption{Ablation study.}
    
\end{table}

We evaluate the effectiveness of each module in our method. Fig.~\ref{fig:pose_refinement} illustrates the principle of each module. In Table.~\ref{Table_ablation:6d_object_velocity_integration}, we report the results of using only local correction, using only 6D object velocity propagation, and our proposed method, which combines the two modules. The results indicate that, with our method, tracking errors (RMSE) for both translation and rotation decrease substantially. Additionally, the tracking trajectories are smoother, as the standard deviation ($\sigma$) is significantly reduced.

\subsubsection{Event optical flow based object velocity}

In the Table.~\ref{Table_ablation:event_op_velocity}
We evaluate the integration of 6D object velocity tracker using event-based optical flow. We compare it with the usage of 6D object velocity calculated from two continuous 6D object poses tracked by local correction. By integrating event-based optical flow based 6D object velocity tracker, the system achieves lower error and better performance compared to using 6D object velocity derived from two continuous poses. This demonstrates the effectiveness of integrating event-based optical flow 6D object velocity tracker.

\subsubsection{Integrate Unscented Kalman filter}

We also evaluate the effectiveness of integrating the Unscented Kalman filter. The filter, applied before the output of the tracked pose, aims to further smooth the pose trajectories. From Table.~\ref{Table_ablation:integrate_ukf}, although the reduction of the standard deviation $\sigma$ of the tracking error for both translation and rotation is modest, it nonetheless represents a positive improvement. Additionally, the tracking error~(RMSE) is also slightly reduced.

\section{DISCUSSION}
FoundationPose demonstrates stable and accurate performance on the test sequences, particularly when there is almost no image blur; however, achieving high-frequency operation requires a high-performance GPU. Additionally, its performance is ultimately constrained to the sensor’s output rate (i.e. 30-60 Hz). In contrast, our method is not affected by image blur, benefiting from the high temporal resolution of event cameras.

We currently do not have an full on-line pipeline that would take advantage of the event-camera for real-time 6-DoF object pose tracking. However, the components of our pipeline were measured independently: velocity estimation (optical flow and KF) takes on average 2 ms, template pose correction takes on average 5 ms, the pose propagation and final UKF take less than 2 ms. Therefore we expect an on-line pipeline should be realized at approximately 110 Hz, while also taking advantage of event camera properties of removed motion blur and high dynamic range.

\textcolor{black}{While our approach relies on an initial pose, in this work we first focus on the tracking algorithm that is most suited to be solved with event cameras due to their nature of measuring appearance change}. For an eventual complete pipeline, a dedicated pose estimator could be introduced (e.g. as in \cite{10342300}) to provide the initial pose and help with failure recovery, however no robust object pose detection networks have been proposed for event cameras as of the time of this publication.

A real-world 6D object pose dataset captured with event cameras, particularly including fast moving objects and annotated with ground truth poses, would be highly valuable for advancing research on 6D object pose tracking or estimation using event cameras. 

\section{CONCLUSIONS}
In this work, we introduce the event-camera-only method using propagation and correction for 6D object pose tracking, which combines event-based optical flow and a template matching approach. By rendering depth using the tracked 6D object pose, our method removes the need for explicit depth measurements required in 6D velocity estimation. We demonstrate the effectiveness of our approach through comparisons on both synthetic and real camera data sequences against existing event-camera pose trackers and state-of-the-art frame-based deep learning methods.






\IEEEtriggercmd{\enlargethispage{-6cm}}
\IEEEtriggeratref{12}
\bibliographystyle{IEEEtran}
\bibliography{biblio}

\end{document}